\documentclass[twocolumn]{IEEEtran}
\usepackage{booktabs}
\usepackage{url}
\usepackage{caption}
\usepackage{graphicx}
\usepackage{color}
\usepackage{array}
\usepackage{here}
\usepackage{float}
\usepackage[utf8]{inputenc}
\usepackage[export]{adjustbox}
\usepackage{multirow}
\usepackage{amsmath}
\usepackage{tcolorbox}
\usepackage{pdfpages}
\usepackage{dblfloatfix}
\usepackage{hyperref} \hypersetup{colorlinks=true, citecolor=black}
\usepackage{subcaption}
\usepackage{multicol}
\usepackage[font=small,labelfont=bf]{caption}
\renewcommand{\baselinestretch}{.995}
\graphicspath{{Images/}}

\hypersetup{
  colorlinks=true,
  linkcolor=blue,
  filecolor=magenta,   
  urlcolor= blue,
}

\begin{document}       

\title{A Generative Model to Synthesize EEG Data for Epileptic Seizure Prediction}
\author{Khansa Rasheed$^1$, Junaid Qadir$^1$, Terence J. O'Brien$^{2}$, Levin Kuhlmann$^{3}$ and Adeel Razi$^{4, 5}$\\\vspace{2mm}
$^1$ Information Technology University (ITU)-Punjab, Lahore, Pakistan\\
$^2$ Department of Neuroscience, Central Clinical School,Monash University, Melbourne, Victoria, Australia \\
$^3$ Faculty of Information Technology, Monash University, Clayton, Victoria, Australia \\
$^4$ Turner Institute for Brain and Mental Health, Monash University, Clayton, Victoria, Australia\\
$^5$ Wellcome Centre for Human Neuroimaging, UCL, London, United Kingdom\\
}

\maketitle

\begin{abstract} Epilepsy, a group of neurological disorders, is characterized by recurrent and unpredictable seizures, which can cause an unexpected loss of awareness for patients resulting in a risk of life-threatening incidents. Prediction of seizure before they occur is therefore vital for bringing normalcy to the lives of patients. For decades researchers have been employing machine learning (ML) techniques based on hand-crafted features for seizure prediction. However, ML methods are too complicated to select the best ML model or the best features. Deep Learning (DL) methods are beneficial in the sense of automatic feature extraction. Lamentably despite decades of research, epilepsy prediction remains an unresolved problem. One of the roadblocks for accurate seizure prediction, among others, is the scarcity of good quality epileptic seizure data. This paper addresses this problem by proposing a deep convolutional generative adversarial network (DCGAN) to generate synthetic EEG data samples. We use two methods to validate the synthesized data namely, one-class SVM and a new proposal which we refer to as convolutional epileptic seizure predictor (CESP). Another objective of our study is to evaluate the performance of well-known deep learning models (e.g., VGG16, VGG19, ResNet50, and Inceptionv3) by training the models on augmented EEG data using transfer learning with average time of 10 min between true prediction and seizure onset. Our results show that the CESP model achieves sensitivity of 78.11\% and 88.21\%, and false prediction rate of 0.27/h and 0.14/h for the training on synthesized and testing on real Epilepsyecosystem and CHB-MIT datasets, respectively. The impressive performance of CESP trained on synthesized data shows that the synthetic data acquired the correlation between features and labels very well. We also show that using the employment of the idea of transfer learning and data augmentation in patient-specific manner provides the highest accuracy with sensitivity of 90.03\% and 0.03 FPR/h which was achieved using Inceptionv3, and that augmenting the data with the samples generated from DCGAN increased the prediction results of our CESP model and Inceptionv3 by 4-5\% as compared to the state-of-the-art traditional augmentation techniques. Finally, we note that the prediction results of CESP achieved by using the augmented data are better than the chance level for both datasets. 

\end{abstract}


\section{Introduction} 

A sudden abnormal, self sustaining electrical discharge in the cerebral networks of the brain is a cause of epileptic seizures (ES). The attack may occur at any time on any day. The unpredictability of duration, seriousness, and time of attack makes it very difficult for patients to perform everyday chores and on occasions can be life-threatening. According to the World Health Organization (WHO), 70 million people around the globe suffer from epilepsy with around one-third of these patients resistant to anti-epileptic medication \cite{world2005promoting}. The early prediction of these attacks before they occur will be helpful for the patients to take precautionary measures and potentially allow the implementation of preventative therapies.


\begin{figure*}[!htb]
\centering
\includegraphics[width=0.99\textwidth]{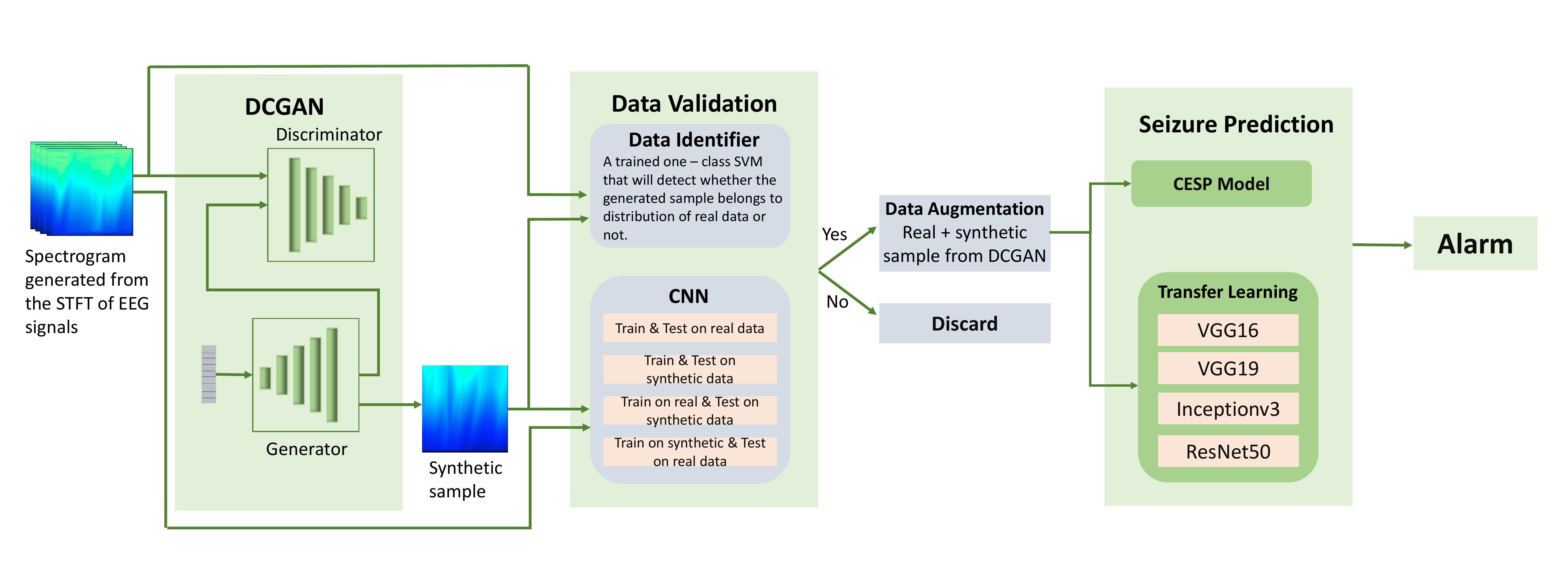}
\caption{Pipeline of the methodology used in the paper.}
\label{fig: pipeline}
\end{figure*}

Electroencephalography (EEG) is used for measuring and monitoring brain activity before, during, and after ES and is widely used to predict seizures. Machine Learning (ML) based prediction algorithm uses the hand-crafted features of EEG from the time-domain, frequency-domain, or time-frequency domain to make predictions. Previously, researchers have evaluated various features---such as Kolmogorov entropy \cite{natarajan2004nonlinear}, largest Lyapunov exponent (LLE) \cite{gotman1985electroencephalographic}, phase synchronization of different EEG channels \cite{iasemidis1990phase}, and correlation density---to perform seizure prediction \cite{le1999anticipating}. In 2014 and 2016, contests of (epileptic) seizure prediction were held by the American Epilepsy Society and Melbourne University. These competitions were open to the EEG feature (for seizure prediction) computing algorithm or ML models trained on the extracted features. However, the submitted algorithms were too complicated to select the best features or the best ML model for seizure prediction \cite{kuhlmann2018epilepsyecosystem}.

Deep learning (DL) algorithms are beneficial in the sense of automatic feature extraction from the data \cite{9139257}. Over the past few years, researchers have applied several DL methods to predict epileptic seizures \cite{truong2017generalised}, \cite{khan2017focal}, \cite{truong2019epileptic}. However, these DL algorithms require an extensive amount of labeled data to produce effective results. Researchers typically use scalp EEG, where signals are collected from the wearable sensors placed on the scalp, or intracranial EEG (iEEG), signals collected by placing the electrode on the exposed surface of the brain through surgery, for the ES prediction. iEEG data gives the high temporal resolution and frequency information of brain activity with high signal to noise ratio as compared to scalp EEG. However, the acquisition of iEEG data is challenging due to surgery and implantation risks.

Despite decades of research in the field of ES prediction, there are still challenges related to the availability of a large amount of labeled EEG data and computationally efficient prediction algorithms. Our major contributions in this paper are as follows:
\begin{enumerate}
  \item We propose a deep convolutional generative adversarial network (DCGAN) to resolve the EEG data scarcity problem. As a proof of principle, we generate synthetic scalp EEG and iEEG data of each patient by training the DCGAN model on the Epilepsyecosystem iEEG data and CHB-MIT scalp EEG data separately in a patient-specific way.
  \item To evaluate the effectiveness of simulated data, we then use two methods: classical ML method of one-class SVM, and secondly a CNN classifier---which we refer to as convolutional epileptic seizure predictor (CESP)---for seizure prediction in the last block of Figure \ref{fig: pipeline}). We train CESP on the real data and test on synthesized data (iEEG data generated from DCGAN). The idea of augmenting the real data with generated data also improved the performance of CESP.
  \item We also evaluate data augmentation with the well-known technique of transfer learning (TL). We trained the popular DL models ResNet50, Inceptionv3, VGG16, and VGG19 on the large amount of already generated synthesized data. After the training, we fine-tuned these pre-trained models on the real data to develop the patient-specific prediction algorithm. 
\end{enumerate}
 
To the best of our knowledge, this is the first study on the generation of synthetic scalp EEG and iEEG data and the use of transfer learning with data augmentation performed by generative methods for seizure prediction. Figure \ref{fig: pipeline} summarizes the method we have developed.

The remaining paper is organized as follows: Section \ref{sec: pre-work} covers the previous work on ES prediction using DL methods. Section \ref{sec:method} provides face validation, i.e., the detailed methodology we propose along with the used dataset. Section \ref{sec:result} provides construct validation, which comprises the results obtained from the proposed methods and comparison of the results with previous work. The paper is concluded in Section \ref{sec:con}.

\section{Related Work}
\label{sec: pre-work}

EEG is a complex and challenging functional brain mapping modality to handle due to the presence of noise and various measurement and physiological artifacts. Pre-processing of EEG data for noise and artifact removal is an involved time-consuming exercise that greatly compromises the utility of online development of EEG based ES prediction solutions \cite{acharya2013automated}. Furthermore, most classical ML-based prediction techniques heavily rely on hand-crafted features \cite{gadhoumi2016seizure}. In contradistinction, the ability of DL-based algorithms to automatically and robustly learn data features from partially or unprocessed EEG data has fueled many new research ideas for accurate and time-efficient ES prediction.

With minimal pre-processing of EEG data, Troung et al. \cite{truong2017generalised} proposed a three-layered convolutional neural network (CNN)-based prediction algorithm for the first time. The auuthors used the spectrograms generated from the short-time Fourier transform (STFT) of the Freiburg hospital iEEG database and the CHBMIT scalp EEG database to train the CNN network. As compared to the interictal or ictal class, the pre-ictal class samples were less, which caused the imbalanced dataset. To overcome this problem of an imbalanced dataset, they also generated preictal samples by sliding a 30sec window over the preictal samples along the time axis. They achieved 89.8\% sensitivity with 0.17 false prediction rate (FPR/h) for 5 min of Seizure Prediction Horizon (SPH). SPH is defined as the time interval between the alarm raised in the anticipation of an impending seizure onset and the actual start of the ictal state \cite{maiwald2004comparison}. Haider et al. \cite{khan2017focal} improved the results by achieving 87.8\% sensitivity and 0.142 FPR for 10 min SPH with the CHBMIT and MSSM databases. They transformed the EEG data into wavelet tensors before training the CNN classifier. Cook et al. \cite{cook2013prediction} demonstrated the feasibility of seizure prediction in clinical setting with the success of implantable EEG recording system. Their work initiated new avenues of research for seizure prediction. Building on the work of Cook et al., Kiral-Kornek et al. \cite{kiral2018epileptic} presented a portable seizure prediction system with tunable parameters according to the patient's need. They transformed iEEG data into spectrograms and used frequency transformed data as an input to the DL model for automatic pre-ictal feature learning. Their prediction system performed at an average sensitivity of 69\% and average time in warning of 27\%, significantly exceeding a comparable random predictor for all patients by 42\%.

Ramy Hussain et al. \cite{hussein2019human} implemented a downsampling technique to reduce the dimension of EEG data but only used fraction of data gathered by Cook et al. They demonstrated that feature extraction is not a viable methodology for reliable ES prediction because the EEG measurements not only vary from patient to patient but also vary for the same patient over time of data acquisition. They applied STFT to transform the EEG data and train a CNN model on this transformed data. They achieved an average of 87.85\% sensitivity and 0.84 area under the curve (AUC). 

In the existing literature, ES prediction is posed as a supervised learning problem. Proposed solutions, which we surveyed above, have been successful however supervised learning requires an abundant amount of labeled data for training process. Expert neurologists and physicians do the manual annotation of the large EEG dataset which is time-consuming and prone to human errors. To address the problem of unavailability of labeled EEG data. Troung et al. \cite{truong2019epileptic} proposed a generative adversarial network (GAN) to perform unsupervised training. They used the spectrogram of STFT of EEG as an input to the GAN. Then, they used the features learned from the discriminator to predict the seizure. To overcome the problem of an imbalanced dataset, they employed the sliding window technique used in their previous work \cite{truong2017generalised}. They measured the area under the receiver operator characteristic (ROC) curve as a performance measure with 5 min SPH and seizure occurrence period (SOP), time interval during which seizure occurrence is expected, of 30 min. They achieved 77.68\% AUC for the CHBMIT scalp EEG data, 75.47\% AUC on the Freiburg hospital data and 65.05\% AUC with the EPILEPSIAE database. 

Data augmentation is a traditional solution to the intricacy of a small dataset. Zhang et al. \cite{zhang2019epilepsy} extracted the time and frequency domain features using wavelet packet decomposition and common spatial pattern (CSP). To address the problem of an imbalanced dataset, they generated artificial pre-ictal samples by randomly combining the segments of original pre-ictal samples and augmented the real data with the generated samples. Then they fed the extracted features to the shallow CNN classifier to predict the seizure. They achieved a sensitivity of 92.2\% and 0.12 FPR/h on the 23 patients of the CHB-MIT dataset. In contrast to these techniques, we used a DL-based method for artificial generation of seizure data.


\section{Methodology}
\label{sec:method}
In this section, we describe the datasets used in the study, pre-processing of these datasets, and then we present the face validation of our proposed methodology.

\subsection{Dataset} 

We are using two datasets for this work: the CHB-MIT dataset \cite{shoeb2009application} and the Epilepsyecosystem dataset \cite{cook2013prediction} (summarized in Table \ref{tab:dataset}). We trained the DCGAN on both datasets separately to generate the samples of scalp EEG and iEEG signals. After the generation of synthetic data, we augmented the real data of both datasets with the synthetic samples. We then employed the idea of transfer learning (TL) on various DL models using augmented Epilepsyecosystem dataset. We also evaluated the performance of the CESP model on both augmented datasets.

The Epilepsyecosystem dataset recorded at St Vincent's Hospital in Melbourne, Australia is used for the experiments. The dataset contains the intracranial EEG (iEEG) signals of three patients (all female). Data of each patient contain signals from 16 electrodes sampled at 400Hz sampling frequency. Data is segmented in 10 min long pre-ictal and interictal intervals. Pre-ictal intervals are selected from the one hour earlier recordings of every seizure with a five-minute seizure horizon while interictal intervals are segmented from randomly selected one-hour recording blocks at least four hours away from any seizure. 

The CHB-MIT dataset consists of the 844h long continuous scalp EEG data of 23 patients with 163 episodes of seizure. Scalp EEG data were collected using 22 electrodes at a sampling rate of 256 Hz. We segmented the interictal periods that are at least 4 hours away before a seizure onset and after a seizure ends. We are interested in anticipating the leading seizures, therefore for two or more consecutive seizures that are less than 30 min apart from each other, we considered these as only one seizure. Moreover, we only considered patients with no more than 10 seizures per day for the ES prediction because it is not very crucial to predict ES for patients that have a seizure every 2 hours on average. With the preceding criteria, there are 13 patients with an adequate amount of data, i.e., have at least 3 leading seizures and 3h of the interictal period.

{\renewcommand{\arraystretch}{1.25}
  \begin{table}[!h]
  \centering
  \caption{Summary of the datasets used in the paper.}
  \label{tab:dataset}
  \begin{tabular}{llllll}
  \hline 
  Dataset & EEG type & \begin{tabular}[c]{@{}l@{}}No. of \\ patients\end{tabular} & \begin{tabular}[c]{@{}l@{}}No. of\\ channels\end{tabular} & \begin{tabular}[c]{@{}l@{}}No. of\\ seizures\end{tabular} & \begin{tabular}[c]{@{}l@{}} Duration \\ average (hr) \end{tabular}\\  \hline \hline
  CHB-MIT & scalp & 13 & 22 & 64 & \begin{tabular}[c]{@{}l@{}} 1-4 \end{tabular} \\ [0.5ex]
  \begin{tabular}[c]{@{}l@{}} Epilepsy-\\-ecosystem\end{tabular} & iEEG & 3 & 16 & 1362 & \begin{tabular}[c]{@{}l@{}} 10608 \end{tabular} \\ \hline
  \end{tabular}
  \end{table}
}


\subsection{Data Preparation}

EEG data was contaminated by the power line noise at 60Hz for the CHB-MIT dataset and 50Hz for the Epilepsyecosystem dataset. We removed the line noise in both datasets using the Butterworth infinite impulse response. After eliminating the line noise, short-time Fourier transform (STFT) is applied to get the spectrogram of EEG signals. We applied STFT on each electrode of the EEG signals of both datasets with 1 min window length with no overlaps. Spectrograms of all electrodes (16 for Epilepsyecosystem and 22 electrodes for CHB-MIT) were concatenated vertically to obtain the final spectrogram.

\subsection{Synthetic Data Generation}

We use a DCGAN to generate synthetic iEEG and scalp EEG data. Here we only describe it for the Epilepsyecosystem dataset. The same description is implemented for the CHB-MIT dataset. The Generator takes a 100 dimensional randomly generated samples from the standard Gaussian distribution of zero mean and standard deviation of one as an input. The input layer is a dense hidden layer. The output dimension of the first hidden layer is 4096 which is reshaped to $4 \times 4 \times 256$. The dense layer is succeeded by 6 de-convolutional layers with a stride size of $2 \times 2$, filter size $ 5 \times 5 $, and the same padding. Number of filters in first de-convolutional layer are 256 and 128 in all other de-convolutional layers. The output of the generator is the same as the spectrograms generated by the STFT ($256 \times 256 \times 3$). 

We configured the discriminator to distinguish the synthetic iEEG data from the real data. The discriminator consists of 4 convolutional layers with 256, 128, 64, and 32 number of filters. The filter size in the convolutional layers is 5 x 5, with a stride of 2x2, and the same padding. While training, the task of the discriminator is to detect whether the spectrograms generated by the Generator are real or fake. The Generator updates its parameters to generate the spectrograms that are not distinctive from real spectrograms \cite{goodfellow2014generative}. 

The equations of discriminator loss $D_{loss}$ and the Generator loss $G_{loss}$ are defined as \cite{goodfellow2014generative}:
\begin{equation}
  D_{loss}=\frac{1}{n}\sum_{j=1}^{n}\left [ \log D(x^{(j)})+\log(1-D(G(z^{(j)}))) \right ]
\end{equation}
\begin{equation}
   G_{loss}=\frac{1}{n}\sum_{j=1}^{n}\log (D(x^{(j)}) )
\end{equation}

where $n$ is our batch size (32), $x$ is the real EEG spectrograms generated from the STFT, and $z$ is a random sample generator from the distribution $\mathcal{N}(0,1)$.

To overcome the problems of overfitting and convergence of discriminator, we configured an early-stopping to have a check on $D_{loss}$ and $G_{loss}$. The early-stopping monitoring stops the training of DCGAN if, over subsequent $k$ training batches, the $D_{loss}$ keeps getting larger than the $G_{loss}$. We used a batch size of 32, $k=15$, Adam as an optimizer for gradient learning with 0.5 value of $\beta_1$, and $1e^{-3}$ learning rate. The value of $G_{loss}$ and $D_{loss}$ achieved the equilibrium point in around 3000 epochs with the early-stopping monitoring. 

We trained the DCGAN on the Epilepsyecosystem iEEG data and the CHB-MIT scalp EEG data separately to obtain the synthetic iEEG and scalp EEG data. To achieve the best results, we trained the DCGAN with three different dataset settings: (i) training on all patients of the Epilepsyecosystem dataset; (ii) training on all patients of the CHB-MIT dataset; and (iii) training of DCGAN only on the pre-ictal class of data of all patients to generate the pre-ictal synthetic samples for data augmentation.

\begin{figure}[!htb]
\centering
\includegraphics[height=9cm]{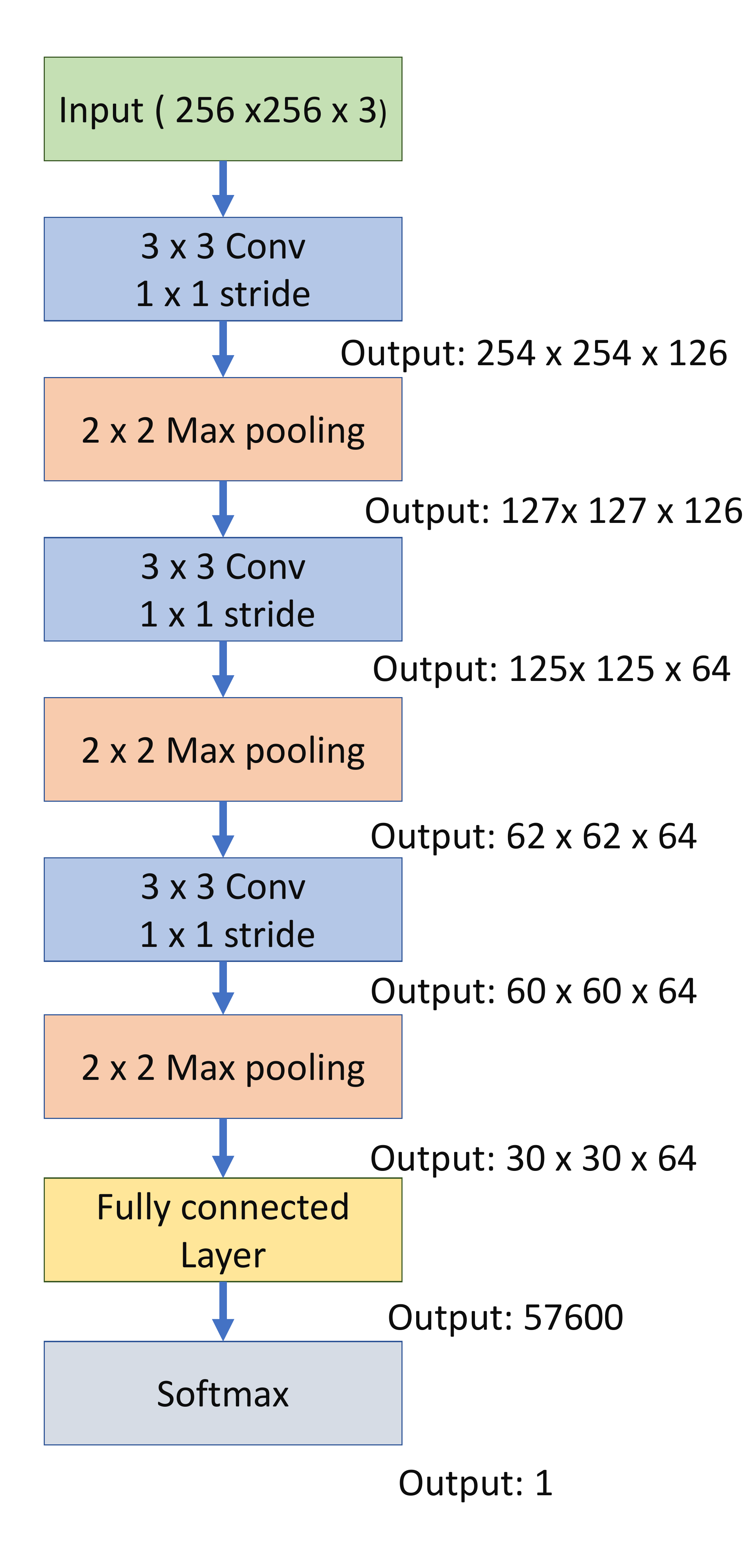}
\caption{Architecture of Convolutional Epileptic Seizure Predictor (CESP).}
\label{fig:CNN-model}
\end{figure}

\subsection{One-class SVM for Data Validation}

Schölkopf et al. \cite{scholkopf2000support} proposed the idea of one-class SVM, which is an extension of a two-class SVM algorithm. One-class SVM is widely utilized to identify the outliers and anomalies in the dataset. A one-class SVM algorithm separates the data from the origin point by a wide margin in the higher dimensional feature space. Then the algorithm computes the surface of a hyperplane, which encloses the anomaly free data (+ve class). The data samples which are out of the hyperplane are outliers/anomalies. The radius of the hyperplane and the number of outliers/anomalies are hyperparameters to select through multiple experiments. 

Let $X$ be the samples of the positive class of dataset such that $ \left \{ x_{i}\in R^{n}, i= 1\cdots l \right \}$, the optimization equation of the algorithm is as follows:
\begin{equation}
\begin{aligned}
& \underset{\xi, \rho, z, b }{\text{minimize}}
& & \frac{1}{2} z^{T}z - \rho +\frac{1}{\nu l}\sum_{i=1}^{l}\xi _{i} \\
& \text{subject to}
& & z^{T}\phi (x_{i})\geq \rho -\xi _{i}, \; \xi _{i}\geq 0.
\end{aligned}
\end{equation}

In the optimization problem above, $\rho $ is the distance of hyperplane from the origin, $ \xi _{i}$ are the hyperparameters, and $\nu \in (0,1]$ selects the fraction of outliers/anomalies outside the hyperplane. The decision function is: 

\[\mathrm{sgn}(\sum_{i}^{l}\beta _{i}K(x_{i},x)-\rho ).\]
We used the Radial Basis Function (RBF) as a kernel function in our experiments:
\[K(x_{i},x_{j}) = e^{-\gamma \left \| x_{i}-x_{j} \right \|^{2}}.\]

\subsection{CESP Model}

CNNs models have been widely used for predicting the seizures successfully in literature \cite{hussein2019human}, \cite{khan2017focal}, due to their ability to learn local dependences of input and the fewer number of trainable parameters due to weight sharing. Based on the state-of-the-art performance of CNN in seizure prediction, we are using the CNN classifier here to evaluate the effectiveness of synthetic data generated from the above-mentioned architecture of DCGAN. The detailed architecture of the proposed model is described in Figure \ref{fig:CNN-model}.

In this work, we used a CNN network that consists of 3 convolutional blocks followed by one fully connected (FC) layer. Each block contains a convolutional layer, a rectified linear unit (ReLu), and a max-pooling layer. The max-pooling technique enables the CNN model to learn temporal or spatially invariant features. The convolutional layers have the filter size of $3 \times 3$, stride $1 \times 1$, $2 \times 2$ size of max-pooling with 126, 64, and 64 number of filters respectively. The FC layers have a sigmoid activation function with output sizes of 32 and 2. We designed this particular architecture to achieve good performance with a simple model. We experimented with a different number of layers of the model and chose the described model of 3 convolutional layers providing good prediction results. To avoid the overfitting for the simple model, we evaluated the training process on the k-fold cross-validation. We used $k=10$ to split the training data into 90\% for training and 10\% for validation. We trained the CESP model for the binary-cross entropy loss on the Adam optimizer with a $1e^{-4}$ learning rate.

\subsection{Transfer Learning}
With the rapidly growing applications of supervised learning in ML, a problem arises when we do not have a sufficient amount of labeled data for training. Transfer learning (TL) deals with this problem by leveraging the already available labeled data of relevant or similar tasks. Since 1993, TL has been used in discriminability-based transfer (DBT) algorithm \cite{pratt1993discriminability}, multi-tasking learning \cite{caruana1997multitask}, cognitive science \cite{pratt1996reuse}, detection of cancer subtypes \cite{hajiramezanali2018bayesian}, text classification \cite{do2006transfer}, and spam filtration \cite{bickel2006ecml}. In another recent work, Bird et al. used TL between EEG signal classification and Electromyographic (EMG) signals \cite{bird2020cross}.

We trained four well-known DL models: VGG16, VGG19, Inceptionv3, and ResNet50 on augmented EEG data. These are well-designed DL models intended to resolve problems of convolutional networks, i.e., vanishing gradient, degradation, long training time, and the large number of trainable parameters. We used the weights of these models trained on the ImageNet dataset as initial weights instead of training from random weights with learning rate of $1e^{-4}$.

\subsection{System Evaluation}
Before evaluating the performance of the prediction algorithm, the SPH, and the Seizure Occurrence Period (SOP) need to be defined. For our work, we are using the definitions established in \cite{maiwald2004comparison}. To make a correct prediction, a seizure must transpire after the SPH and within the SOP. A false alarm will be raised if the prediction algorithm gives a positive signal (seizure is going to occur) but there is no seizure during the SOP. For the best clinical use, the SPH must be long enough to give a patient sufficient time to take precautionary measures after the alarm is raised. We use sensitivity, FPR/h, specificity, and accuracy with SPH of 10 min and SOP of 30 min. For the Epilepsyecosystem the value of SPH is fixed only for the training dataset. As no information about the segmentation timing relative to seizures is provided for the test set, we cannot determine the exact value of SPH, we only have the information that we are 65 to 5 minutes away from the seizure.

We also compared the performance of CESP model trained on the augmented data with a random predictor. Using the method proposed by Schelter et al. \cite{schelter2006testing}, we computed the probability of alarm generation in the duration of SOP for a given value of FPR:
\begin{equation}
\begin{aligned}
\textit{P}\approx 1-e^{-FPR\times SOP}.
\end{aligned}
\end{equation}
Then the probability to predict at least \textit{n} out of \textit{N} independent seizure events at random can be calculated using the following equation:
\begin{equation}
\begin{aligned}
\textit{p}= \sum_{k\geq n}\binom{N}{k}\textit{P}^{k}(1-\textit{P})^{N-k}.
\end{aligned}
\end{equation}
Using the FPR value of each patient and the number of true predictions using CESP (\textit{n}), we computed the p-value for each patient. We can infer that for the significance level of $\alpha$ = 0.05, our approach performed better than a chance level predictor.

\section{Results}
\label{sec:result}
To generate the synthetic data, we trained the DCGAN on the iEEG data of Epilepsyecosystem and the scalp EEG data of CHB-MIT datasets. In this section, we test the effectiveness of synthetic data using the one-class SVM and CESP. The selection of samples of generated data is based on the results of one-class SVM and CESP model. These selected samples were then used for data augmentation for ES prediction. For the CESP we applied different combinations of real and synthetic data while testing and training. Detailed results are described in Table \ref{tab:MIT-table}, \ref{tab:ecosys-tab}.

\begin{figure}[!h]
   \centering
   \begin{subfigure}[h]{0.5\textwidth}
     \centering
     \includegraphics[width=0.8\textwidth,height=3cm]{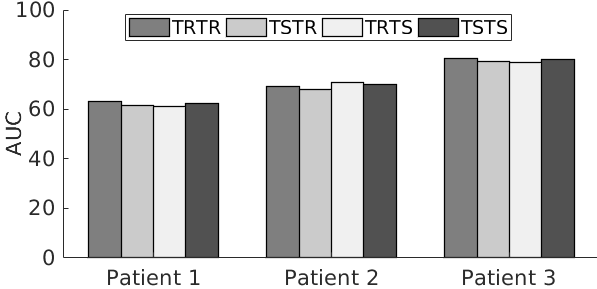}
     \caption{Epilepsyecosystem}
     \label{fig: a}
   \end{subfigure}
   \begin{subfigure}[h]{0.5\textwidth}
     \centering
     \includegraphics[width=0.78\textwidth,height=3cm]{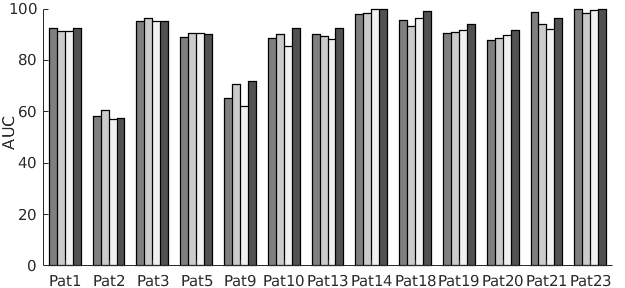}
     \caption{CHB-MIT}
     \label{fig: b}
   \end{subfigure}
  \caption{Seizure prediction performance (AUC) of CESP using different combinations of real (iEEG data of Epilepsyecosystem, scalp EEG data of CHB-MIT) and synthetic (generated iEEG data from DCGAN trained on Epilepsyecosystem dataset, generated scalp EEG data from DCGAN trained on the CHB-MIT dataset) for testing and training. \textbf{Legends}: \textbf{TRTR} = train \& test on real data, \textbf{TSTR} = train on synthesized data \& test on real data, \textbf{TRTS} = train on real data \& test on synthesized data, \textbf{TSTS} = train \& test on synthesized data.}
  \label{fig:testing}
\end{figure}

\begin{table*}[!h]
\centering
\caption{Validation of synthesized pre-ictal samples of scalp EEG data using different combinations of synthesized and real data for training and testing on CESP. Synthesized data generated from the \textbf{DCGAN trained on the CHB-MIT dataset.} The comparison of results of i) test \& train on real data (TRTR) and ii) train on synthetic data \& test on real data (TSTR), shows that the synthetic data has fully captured the correlation between the features of data and the labels. \textbf{Legends}: \textbf{TRTR} = train \& test on real data, \textbf{TSTR} = train on synthesized data \& test on real data, \textbf{TRTS} = train on real data \& test on synthesized data, \textbf{TSTS} = train \& test on synthesized data, \textbf{Sen} = Sensitivity, \textbf{Spec} = Specificity, \textbf{Acc} = Accuracy.}
\label{tab:MIT-table}
\begin{tabular}{|c|c|c|c|l|c|c|c|l|c|c|c|l|c|c|c|l|} 
\hline
\multirow{2}{*}{\begin{tabular}[c]{@{}c@{}} \textbf{Patie-}\\\textbf{-nts} \end{tabular}} & \multicolumn{4}{c|}{\textbf{TRTR} } & \multicolumn{4}{c|}{\textbf{TSTR} } & \multicolumn{4}{c|}{\textbf{TRTS} } & \multicolumn{4}{c|}{\textbf{TSTS} } \\ 
\cline{2-17} & \begin{tabular}[c]{@{}c@{}}\textbf{Sen}\\\textbf{(\%)} \end{tabular} & \multicolumn{1}{l|}{\textbf{FPR/h} } & \multicolumn{1}{l|}{\begin{tabular}[c]{@{}l@{}}\textbf{Spec}\\\textbf{~(\%)} \end{tabular}} & \begin{tabular}[c]{@{}l@{}}\textbf{Acc}\\\textbf{(\%)}\end{tabular} & \multicolumn{1}{l|}{\begin{tabular}[c]{@{}l@{}}\textbf{Sen}\\\textbf{~(\%)} \end{tabular}} & \multicolumn{1}{l|}{\textbf{FPR/h} } & \multicolumn{1}{l|}{\begin{tabular}[c]{@{}l@{}}\textbf{Spec}\\\textbf{~(\%)} \end{tabular}} & \begin{tabular}[c]{@{}l@{}}\textbf{Acc}\\\textbf{(\%)}\end{tabular} & \multicolumn{1}{l|}{\begin{tabular}[c]{@{}l@{}}\textbf{Sen}\\\textbf{~(\%)} \end{tabular}} & \multicolumn{1}{l|}{\textbf{FPR/h} } & \multicolumn{1}{l|}{\begin{tabular}[c]{@{}l@{}}\textbf{Spec}\\\textbf{~\textbf{(\%})} \end{tabular}} & \begin{tabular}[c]{@{}l@{}}\textbf{Acc}\\\textbf{(\%)}\end{tabular} & \multicolumn{1}{l|}{\begin{tabular}[c]{@{}l@{}}\textbf{Sen}\\\textbf{~(\%)} \end{tabular}} & \multicolumn{1}{l|}{\textbf{FPR/h} } & \multicolumn{1}{l|}{\begin{tabular}[c]{@{}l@{}}\textbf{Spec}\\\textbf{~(\%)} \end{tabular}} & \begin{tabular}[c]{@{}l@{}}\textbf{Acc}\\\textbf{(\%)}\end{tabular} \\ 
\hline
Pat1 & 91.59 & 0.19  & 81 & 92.03 & 88.15  & 0.24  & 76 & 90.18 & 88.09 & 0.25 & 75& 90.85 & 90.58 & 0.18  & 82 & 91.57 \\ 
\hline
Pat2 & 75.2 & 0.01 & 99  & 89.24 & 70.59& 0.04  & 96 & 85.49  & 68.13 & 0.01 & 99 & 84.08 & 71.23 & 0.02 & 98 & 88.93 \\ 
\hline
Pat3 & 95.11 & 0.14  & 86 & 96.11 & 96.36 & 0.17 & 83 & \multicolumn{1}{c|}{96} & 95.21 & 0.15 & 85 & 95.55  & 95 & 0.14 & 86 & 94.89 \\ 
\hline
Pat5  & 89.1  & 0.15 & 85 & 88.36 & 78.59 & 0.19 & 81 & 83.19 & 88.64  & 0.18 & 82  & 86.52  & 90.26 & 0.16 & 84  & 89.91 \\ 
\hline
Pat9 & 65.23 & 0.01 & 99 & 78.51 & 70.51 & 0.02 & 98 & 80.26 & 62 & 0.00 & 100 & 89.97 & 71.87 & 0.00 & 100  & 89.99 \\ 
\hline
Pat10 & 88.56 & 0.16 & 84 & 89.91 & 90.23 & 0.15 & 85  & 89.99 & 85.27  & 0.13 & 87  & 88.92  & 92.58 & 0.12 & 88 & 90.56 \\ 
\hline
Pat13 & 90.14 & 0.22 & 78 & 91.15 & 89.36 & 0.19 & 81 & 91.05 & 88.06  & 0.20 & 80 & 90.18 & 92.41 & 0.18 & 82  & 91.07 \\ 
\hline
Pat14 & 99.01 & 0.15 & 85 & 96.34 & 98.24 & 0.16 & 84 & 95.74 & 100 & 0.12& 88 & 98.43 & 100 & 0.11& 89 & 97.87 \\ 
\hline
Pat18 & 95.41 & 0.21 & 79 & 89.37 & 93.36  & 0.25 & 75 & 88.43 & 96.28 & 0.24 & 76 & 92.09  & 98.99 & 0.20   & 80 & 91.99\\ 
\hline
Pat19 & 90.38 & 0.29 & 71 & 92.31 & 91 & 0.30  & 70  & 91.50 & 91.45 & 0.31 & 69 & 90.83 & 93.89 & 0.25  & 75 & 91.38 \\ 
\hline
Pat20 & 87.58 & 0.18  & 82 & 90.84 & 88.45  & 0.14 & 86 & 89.09 & 89.63 & 0.16 & 84  & 90.44 & 91.56   & 0.15 & 85 & 90.59\\ 
\hline
Pat21 & 93.57 & 0.00 & 100 & 94.83 & 93.79 & 0.02 & 98 & 92.58 & 92.03 & 0.00 & 100 & 95.14 & 96.32 & 0.01    & 99 & 96.46 \\ 
\hline
Pat23 & 100 & 0.05 & 95 & 97.62 & 98.1 & 0.03 & 97 & 97.32 & 99.56 & 0.07 & 93 & 92.21 & 100 & 0.02 & 98    & 91.17 \\ 
\hline
Average & 89.28 & 0.13 & 87 & 91.24 & 88.21 & 0.139 & 86 & 90.03 & 88.02 & 0.14 & 86 & 91.17 & 90.9  & 0.11   & 89 & 92.02 \\
\hline
\end{tabular}
\end{table*}

\begin{table*}[!h]
\centering
\caption{Validation of synthesized pre-ictal samples of iEEG data using different combinations of synthesized and real data for training and testing on CESP. Synthesized data generated from the \textbf{DCGAN trained on Epilepsyecosystem dataset.} The comparison of results of: i) test \& train on real data (TRTR) and ii) train on synthetic data \& test on real data (TSTR), shows that the synthetic data has fully captured the correlation between the features of data and the labels.}
\label{tab:ecosys-tab}
\begin{tabular}{|l|c|c|c|c|c|c|c|c|c|c|c|c|} 
\hline
\multirow{2}{*}{\begin{tabular}[c]{@{}l@{}}\textbf{Data}\end{tabular}} & \multicolumn{4}{c|}{\textbf{Patient 1} } & \multicolumn{4}{c|}{\textbf{Patient 2} } & \multicolumn{4}{c|}{\textbf{Patient 3} }\\ 
\cline{2-13} & \multicolumn{1}{l|}{\begin{tabular}[c]{@{}l@{}}Accuracy\\(\%) \end{tabular}} & \multicolumn{1}{l|}{\begin{tabular}[c]{@{}l@{}}Sensitivity\\(\%) \end{tabular}} & \multicolumn{1}{l|}{FPR/h} & \begin{tabular}[c]{@{}l@{}}Specificity\\(\%)\end{tabular} & \multicolumn{1}{l|}{\begin{tabular}[c]{@{}l@{}}Accuracy\\(\%) \end{tabular}} & \multicolumn{1}{l|}{\begin{tabular}[c]{@{}l@{}}Sensitivity\\(\%) \end{tabular}} & \multicolumn{1}{l|}{FPR/h} & \begin{tabular}[c]{@{}l@{}}Specificity\\(\%)\end{tabular} & \multicolumn{1}{l|}{\begin{tabular}[c]{@{}l@{}}Accuracy\\(\%) \end{tabular}} & \multicolumn{1}{l|}{\begin{tabular}[c]{@{}l@{}}Sensitivity\\(\%) \end{tabular}} & \multicolumn{1}{l|}{FPR/h} & \begin{tabular}[c]{@{}l@{}}Specificity\\(\%)\end{tabular} \\ 
\hline
TRTR & 72.59 & 78.01 & 0.4 & 60  & 73.00  & 82.16  & 0.31 & 69  & 69.81  & 75.00  & 0.22 & 78 \\ 
\hline
TSTR & 70.12  & 76.59 & 0.38 & 62 & 74.21  & 81.46 & 0.25 & 75 & 71.24 & 76.30 & 0.20  & 80 \\ 
\hline
TRTS & 68.98 & 75.30 & 0.39 & 61 & 71.18  & 79.23  & 0.28 & 72 & 70.00  & 72.39 & 0.23 & 77  \\ 
\hline
TSTS & 72.00 & 79.18 & 0.39 & 61 & 72.99 & 82.00  & 0.22 & 78  & 70.98 & 75.11 & 0.21 & 79 \\
\hline
\end{tabular}
\end{table*}

\begin{figure*}[!h]
   \centering
   \begin{subfigure}[h]{0.8\textwidth}
     \centering
     \includegraphics[height=5cm]{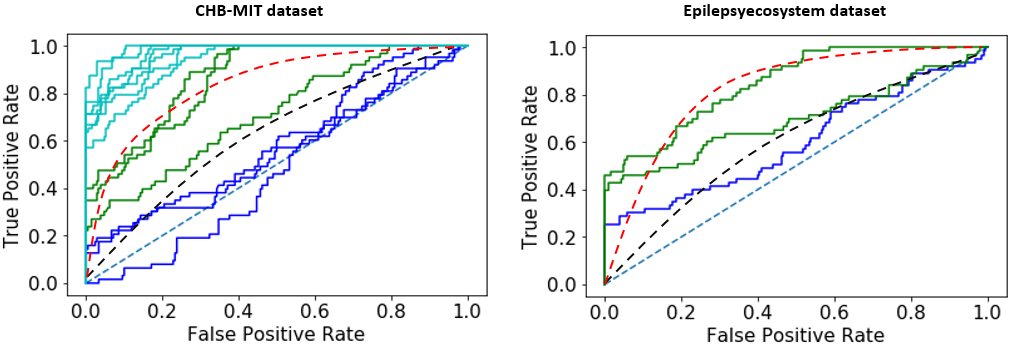}
     \caption{Receiver operating characteristics (ROC) curves without data augmentation}
     \label{fig: a-augment}
   \end{subfigure}
   \par\bigskip
   \begin{subfigure}[h]{0.8\textwidth}
     \centering
     \includegraphics[height=5cm]{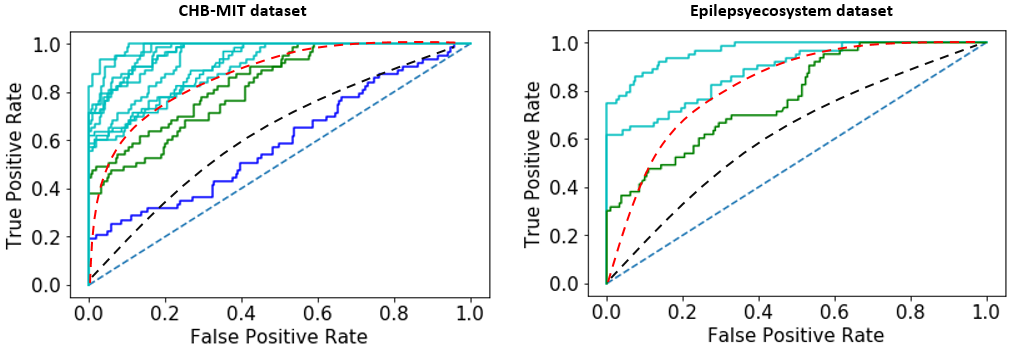}
     \caption{Receiver operating characteristics (ROC) curves with data augmentation}
     \label{fig: b-augment}
   \end{subfigure}
  \caption{Receiver operating characteristics (ROC) curves of ES prediction performance testing for two datasets with and without augmentation: (a) without augmenting the datasets (b) with data augmentation using the synthetic samples from DCGAN. In these curves, each line represents a patient. Above the black dash line: good prediction performance; above the red dash line; very good prediction performance (adapted from \cite{kuhlmann2018seizure}).}
  \label{fig:AUC}
\end{figure*}

\begin{figure}[!h]
   \centering
   \begin{subfigure}[h]{0.4\textwidth}
     \centering
     \includegraphics[width=0.9\textwidth]{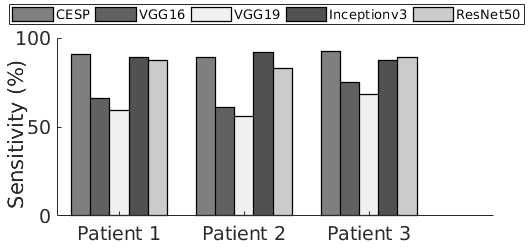}
     \caption{}
     \label{fig: a-TL}
   \end{subfigure}
   \begin{subfigure}[h]{0.4\textwidth}
     \centering
     \includegraphics[width=0.9\textwidth]{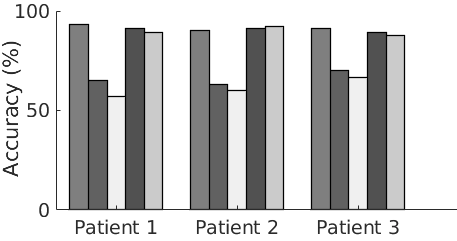}
     \caption{}
     \label{fig: b-TL}
   \end{subfigure}
   \begin{subfigure}[h]{0.4\textwidth}
     \centering
     \includegraphics[width=0.9\textwidth]{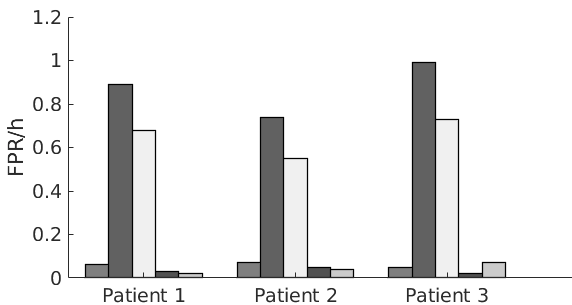}
     \caption{}
     \label{fig: c-TL} 
   \end{subfigure}
   \begin{subfigure}[h]{0.4\textwidth}
     \centering
     \includegraphics[width=0.9\textwidth]{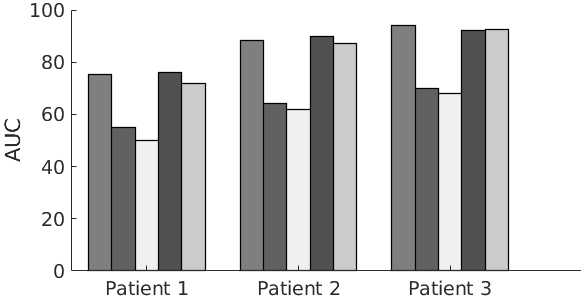}
     \caption{}
     \label{fig: d-TL} 
   \end{subfigure}
  \caption{Comparison of performance of DL models trained on augmented \textbf{Epilepsyecosystem dataset} using transfer learning. }
  \label{fig:TL_results}
\end{figure}

For the selection of correct generated samples, we trained the one-class SVM algorithm on the datasets of real EEG signals separately. We considered that the real data is a positive class and the anomalous data is a negative class. The algorithm learns the distribution of real EEG data and classifies the generated data sample in a positive class or negative class. After training one-class SVM on real data, we tested it for the synthesized samples. We selected those synthesized samples which belonged to the positive class and discarded the samples that belonged to the negative class (see Figure \ref{fig: pipeline}).

\begin{table}[!htb]
  \caption{Comparison of CESP performance trained on the augmented data with the random predictor.}
  \label{tab:4}
  \begin{subtable}{.5\linewidth}
   \centering
    \caption{For CHB-MIT dataset}
    \label{tab:4a}
    \begin{tabular}{|l|c|}
    \hline
    \textbf{Patient} & \multicolumn{1}{l|}{\textbf{\textit{p}-value}} \\ \hline
    Pat1 & \textless{}.001 \\ \hline
    Pat2 & .029 \\ \hline
    Pat3 & \textless{}.001 \\ \hline
    Pat5 & \textless{}.001 \\ \hline
    Pat9 & \textless{}.001 \\ \hline
    Pat10 & .001 \\ \hline
    Pat13 & \textless{}.001 \\ \hline
    Pat14 & .002 \\ \hline
    Pat18 & \textless{}.001 \\ \hline
    Pat19 & 0.03 \\ \hline
    Pat20 & \textless{}.001 \\ \hline
    Pat21 & \textless{}.001 \\ \hline
    Pat23 & \textless{}.001 \\ \hline
    \end{tabular}
  \end{subtable}%
  \begin{subtable}{.5\linewidth}
   \centering
    \caption{For Epilepsyecosystem dataset.}
    \label{tab:4b}
    \begin{tabular}{|l|l|}
    \hline
    \textbf{Patient} & \textbf{p-value} \\ \hline
    Patient1 & .008 \\ \hline
    Patient2 & \textless{}.001 \\ \hline
    Patient3 & .0035 \\ \hline
    \end{tabular}
  \end{subtable} 
\end{table}

\begin{table*}[!h]
\centering
\caption{Comparison of our work with previous works.}
\label{tab:augm-comp}
\begin{tabular}{|l|l|l|l|l|l|l|l|} 
\hline
\textbf{Year} & \textbf{Authors}& \begin{tabular}[c]{@{}l@{}}\textbf{No. of seizure}\\\textbf{in original} \\\textbf{dataset } \end{tabular} &\multicolumn{1}{c|}{\textbf{Features} }& \multicolumn{1}{c|}{\textbf{Classifier} } & \begin{tabular}[c]{@{}l@{}}\textbf{Sensitivity}\\\textbf{ (\%)} \end{tabular} & \textbf{FPR/h} & \textbf{Augmentation technique} \\ 
\hline
\multicolumn{8}{|l|}{\textbf{CHB-MIT}} \\ 
\hline
2017& \begin{tabular}[c]{@{}l@{}}Truong et \\ al. \cite{truong2017generalised} \end{tabular}  & 64 & STFT & CNN & 81.2 & 0.16 & Windowing\\                
\hline
2020& \begin{tabular}[c]{@{}l@{}}Zhang et \\ al. \cite{zhang2019epilepsy} \end{tabular}   & 156 & \begin{tabular}[c]{@{}l@{}}Common spatial\\ features \end{tabular} & CNN & 92 & 0.12 & \begin{tabular}[c]{@{}l@{}} Division of a data sample into 3\\ smaller segments and generation \\of artificial sample from random \\concatenation of these segments \end{tabular} \\ 
\hline
\textbf{2020}& \textbf{This work} & \textbf{64} &\textbf{ STFT} & \textbf{CESP} & \textbf{96} & \textbf{0.05} & \textbf{\begin{tabular}[c]{@{}l@{}}Samples generated form \\ DCGAN \end{tabular}}\\
\hline
\multicolumn{8}{|l|}{\textbf{Epilepsyecosystem}}  \\
\hline
2020 & \begin{tabular}[c]{@{}l@{}}Stojanovic\\ et al. \cite{stojanovic2020predicting} \end{tabular} & 692 & \begin{tabular}[c]{@{}l@{}}Non-negative\\ matrix factorization\end{tabular} & SVM & 69 &0.76 & Synthetic monitory over-sampling \\ 
\hline
2020 & \begin{tabular}[c]{@{}l@{}}Ramy et \\ al. \cite{hussein2020epileptic} \end{tabular}   & 1326  & \begin{tabular}[c]{@{}l@{}}Continuous wavelet\\ transform \end{tabular}   & \begin{tabular}[c]{@{}l@{}}Semi-Dilated\\ CNN \end{tabular} & 89.52  & N/M  & \begin{tabular}[c]{@{}l@{}} Division of data into \\ smaller segments \end{tabular}\\ 
\hline
2020 & This work & 1326 & STFT  & CESP & 92.87  & 0.15 & \begin{tabular}[c]{@{}l@{}} Samples generated form \\ DCGAN \end{tabular}  \\
\hline
\end{tabular}
\end{table*}

To further validate the selected samples, we performed the testing and training of the CESP model for four data combinations. We train and test the model on real data (TRTR) to check and compare the performance of the model with generated data. Then, we test the model trained on real data for samples of synthetic data (TRTS). We also trained the model on synthetic data and evaluated the performance on the real data (TSTR) to validate the selected samples of synthetic data. The results of these experiments for the Epilepsyecosystem and CHB-MIT datasets are provided in Table \ref{tab:MIT-table} and \ref{tab:ecosys-tab} respectively. Figure \ref{fig:testing} depicts the AUC of ES prediction results for multiple scenarios of testing and training. For the Epilepsyecosystem data, we achieved an average sensitivity of 78.39\% for TRTR and 77.56\% for TSTR. It shows that the generated samples selected from the one-class SVM are correct. Similar is the case for the CHB-MIT dataset, we achieved an average of 89.02\% sensitivity for TRTR and 88.21\% sensitivity for TSTR.

We experimented with the training of the CESP model on the $5\times$ and $3\times$ augmented Epilepsyecosystem and the CHB-MIT dataset respectively. Compared to the results achieved by using unaugmented data, data augmentation using DCGAN increased the sensitivity $\sim15\%$ and AUC $\sim10\%$ for Epilepsyecosystem dataset. For the CHB-MIT dataset, AUC increased $\sim6\%$ for augmented data. Figure \ref{fig:AUC} demonstrates the overall ES prediction performance for two datasets with and without augmentation. Table \ref{tab:4} shows the statistical comparison of CESP model trained on the augmented data using the proposed approach of data generation with the chance level predictor. Results indicate that the performance of CESP model for both datasets is better than the chance level predictor. We also compared the results of our data augmentation approach with the previous works in Table \ref{tab:augm-comp}. The comparison shows a significant increase in the prediction results by augmenting the data with synthetic samples for both datasets.


With the availability of augmented data, we evaluated the performance of four widely used DL models: VGG16, VGG19, Inceptionv3, and ResNet50. These models are used for image classification and weights of these models trained on the ImageNet dataset are available in Keras. We twice trained these models on augmented iEEG data with the pre-trained weights as initial weights. First we trained all models on the augmented data of all patients and then we fine-tuned the models in a patient-specific manner. Figure \ref{fig:TL_results} depicts the results of seizure prediction with the models trained on Epilepsyecosystem augmented data. The performance of Inceptionv3 and ResNet50 is considerable as compared to the VGG16 and VGG19. VGG16 and VGG19 have more trainable parameters and required more training time. VGG16 and VGG19 overfit to the training data after some time of training which leads to the poor testing performance. However, the performance of other two models is good enough to use and explore the idea of training these models on adequate amount of data for seizure prediction in future. 

Table \ref{tab:p-value} shows the comparison of AUCs of TL models with the CESP. \textit{p}-values for the VGG16, VGG19, and ResNet50 indicate a significant difference between the performance and AUC curves of prediction models. The performance of Inceptionv3 is best among the TL algorithms and approximately equal to the performance of the CESP model. However, the advantage of the CESP model is the low computational cost and complexity. 

\begin{table}[h]
\centering
\caption{Statistical comparison (\textit{p}-values) of prediction models. \textit{p}-values are derived from the single-tailed Hanley-McNeil test for comparing AUCs.}
\label{tab:p-value}
\begin{tabular}{ccccc}
\hline
\textbf{Patients} & \textbf{\begin{tabular}[c]{@{}c@{}}CESP\\  VS\\ VGG16\end{tabular}} & \textbf{\begin{tabular}[c]{@{}c@{}}CESP\\  VS\\ VGG19\end{tabular}} & \textbf{\begin{tabular}[c]{@{}c@{}}CESP\\ VS\\ Inceptionv3\end{tabular}} & \textbf{\begin{tabular}[c]{@{}c@{}}CESP\\ VS\\ ResNet50\end{tabular}} \\ \hline
Patient1 & \textless{}.000001* & \textless{}.000001* & .046 & .000021* \\ \hline
Patient2 & \textless{}.000001* & \textless{}.000001* & .085 & .005* \\ \hline
Patient3 & \textless{}.000001* & \textless{}.000001* & .059 & .0025* \\ \hline
\end{tabular}\\
* indicates significant p-values after penalizing for correction for multiple comparisons using Bonferroni for each patient at the level $\alpha$ = 0.05/4.
\end{table}



\section{Discussion}

This work aimed to address the scarcity problem of good quality EEG data for ES prediction. With the advancement of DL techniques, high-quality artificial data generation is now possible. Deep generative models trained in an adversarial manner can simulate complex data distributions. In this paper, we presented a DL based generated model (DCGAN) that can generate the artificial EEG samples of patients. After measuring the quality of data using a traditional one-class SVM model and four different tests and training experiments, we augmented the real data with synthetic pre-ictal samples. We then trained CESP model on the augmented data and compared the performance of the model for ES prediction with previous works that used traditional augmentation techniques, i.e., SMOTE, moving windows, and data sampling. The comparison shows that the prediction performance using synthetic pre-ictal samples increased for both datasets. Figure \ref{fig:AUC} demonstrated that the ES prediction performance for all patients of both datasets increased than a chance level predictor by using the augmented data.

In contrary to previously used augmentation techniques, our technique generates artificial samples of data, which is a solution to medical data sharing problems. Besides data acquisition difficulties, medical data sharing comes with the privacy-preserving issues. Researchers and hospitals cannot use the data without the permission of patients and ethical approval \cite{van2014systematic}. The synthetic data is not only used to augment the real data for performance improvement but also can be shared with researchers without privacy issues. 

Previous works \cite{stojanovic2020predicting}, \cite{hussein2020epileptic} using synthetic monitoring over-sampling (SMOTE) and data division into smaller segments techniques for augmentation achieved the sensitivity of 69\% and 89.52\% for the Epilepsyecosystem dataset. Authors in \cite{stojanovic2020predicting} achieved the results by extracting 12 non-negative matrix features and used 692 original pre-ictal samples while we trained DCGAN on the 1362 original samples without any feature extraction. The generalization of the prediction technique is a major pitfall of many research works. The generalization of the model requires the prediction performance for different datasets. The work in \cite{zhang2019epilepsy} divided the original pre-ictal samples into 3 smaller samples and concatenated these samples with random selection to generate new data samples that belonged to the distribution of real data. Their augmentation technique provided significant results. However, they tested the augmentation methods for only one dataset which had a small number of seizures per patient. They also employed feature engineering techniques on the data before feeding the data to CNN. Our data generation is more generalized and applicable to both iEEG and scalp EEG data.

To date, researchers and developers are applying traditional ML and DL techniques for ES prediction. However, with the availability of computational resources and an ample amount of data, TL is an emerging technique to implement for problem-solving. In this paper, we presented the use of famous DL models, i.e., VGG16, VGG19, Inceptionv3, and ResNet50, for the first time to predict seizure. With the availability of augmented data, the experiments performed on these models showed significant results. More precisely, the performance results of Inceptionv3 and ResNet50 were accurate enough to use these pre-trained models for future works, e.g., use the model for new patients with fine tunning, use the model for prediction of other diseases with EEG signals, or use these pre-trained models for extracting significant features of EEG data and make predictions based on extracted features. Besides the promising results of Inceptionv3 and ResNet50, due to the computational complexities of these models, a clinically implantable device will not be an appropriate idea for these models. CESP model has less computational cost as compared to these two models. So, another future direction of TL work can be the reduction of complexities of these models while preserving the performance efficiency, i.e., we can utilize the features extracted from selected layers of models and predict seizures based on these features.

For the purpose of simplicity and ease of comparison, the research community stated the ES prediction problem as a binary classification problem. We have performed all the experiments with the binary classification assumption. However, for the clinical application of these experiments, the formulation of the problem will be complex because seizure highly depends on patient-specific characteristics, i.e., seizure type, age and gender of the patient, and the medication that the patient was taking during data acquisition. For evaluating the prediction models reliably, we need continuous seizure data because the pitfall of AUC performance metric is that it is typically calculated for the balanced dataset (same number of pre-ictal and interictal samples). However, the actual data contained more inter-ictal periods as compared to pre-ictal events. 

 Unavailability of annotated data, privacy-preserving issues, and the ethical problems regarding private data sharing come with the promising results of ML and DL models. Artificial data generation is one solution to these problems. However, the time-series, i.e., EEG data for seizure prediction, contains information that appears many hours ago from the seizure event but is as useful as the information appear one minute before the seizure. That is why the generation of continuous data is one of the future extension of our work. With the significant results of synthesized data samples for seizure prediction, the generation of artificial patient's data is also possible. In this way, researchers can work with an ample amount of data of various patients to address the generalization problem.

\section{Conclusions}
\label{sec:con}

The main aim of ES predictions research is to provide an accurate seizure warning system to patients to take precautionary measures ahead of seizure onset. However, such a solution is not yet available due to scarcity of suitable amount of seizure EEG data. In this paper, we proposed a deep convolutional generative adversarial network (DCGAN) model to overcome the hurdle of the unavailability of an extensive amount of EEG data. The proposed DCGAN model showed good generalization for the generation of both iEEG and scalp EEG data. Moreover, a convolutional epileptic seizure predictor (CESP), was proposed to validate the synthetic data, is also generalized to work with both types of EEG data. To measure the quality of synthetic data, we employed one-class SVM and training and testing of the CESP model with four combinations of real and synthetic data. The CESP model produced sensitivity of 78.11\%, 88.21\% and FPR/h of 0.27, 0.14 for training on synthesized and testing on real Epilepsyecosystem and CHB-MIT datasets respectively. These results are higher than the training and testing of the CESP model on real data. This shows that the synthetic samples fully captured the relationship between the features of data and the labels of pre-ictal samples. We also evaluated the performance of CESP, VGG16, VGG19, Inceptionv3, and ResNet50 on the augmented dataset using the concept of transfer learning (TL). Using the TL on augmented data, we showed that the Inceptionv3 performed very well with highest accuracy of 90.03\% and 89.50\% sensitivity. With these significant results, using TL, we can further explore this novel idea of employing TL techniques for ES prediction in the future work.

\section{Acknowledgments}
Adeel Razi is funded by the Australian Research Council (Refs: DE170100128 and DP200100757) and Australian National Health and Medical Research Council Investigator Grant (Ref: 1194910).

\bibliographystyle{IEEEtran}

\end{document}